\documentclass[conference,a4paper]{IEEEtran}
\IEEEoverridecommandlockouts

\usepackage[hidelinks]{hyperref}
\usepackage[cmex10]{amsmath}
\usepackage{amssymb,amsfonts}
\usepackage{dblfloatfix}

\usepackage[ruled,vlined]{algorithm2e}
\usepackage{graphicx}

\usepackage{booktabs}
\usepackage{siunitx}
\usepackage[numbers,compress]{natbib}
\usepackage{texnames}
\usepackage{bm,bbm}
\usepackage{orcidlink}
\usepackage{tabularx}
\usepackage{cuted}
\usepackage{caption}
\usepackage{subcaption}

\newcolumntype{C}{>{\centering\arraybackslash}X}

\begin{document}

\title{
    \begin{minipage}{0.10\textwidth}
    \includegraphics[width=\textwidth]{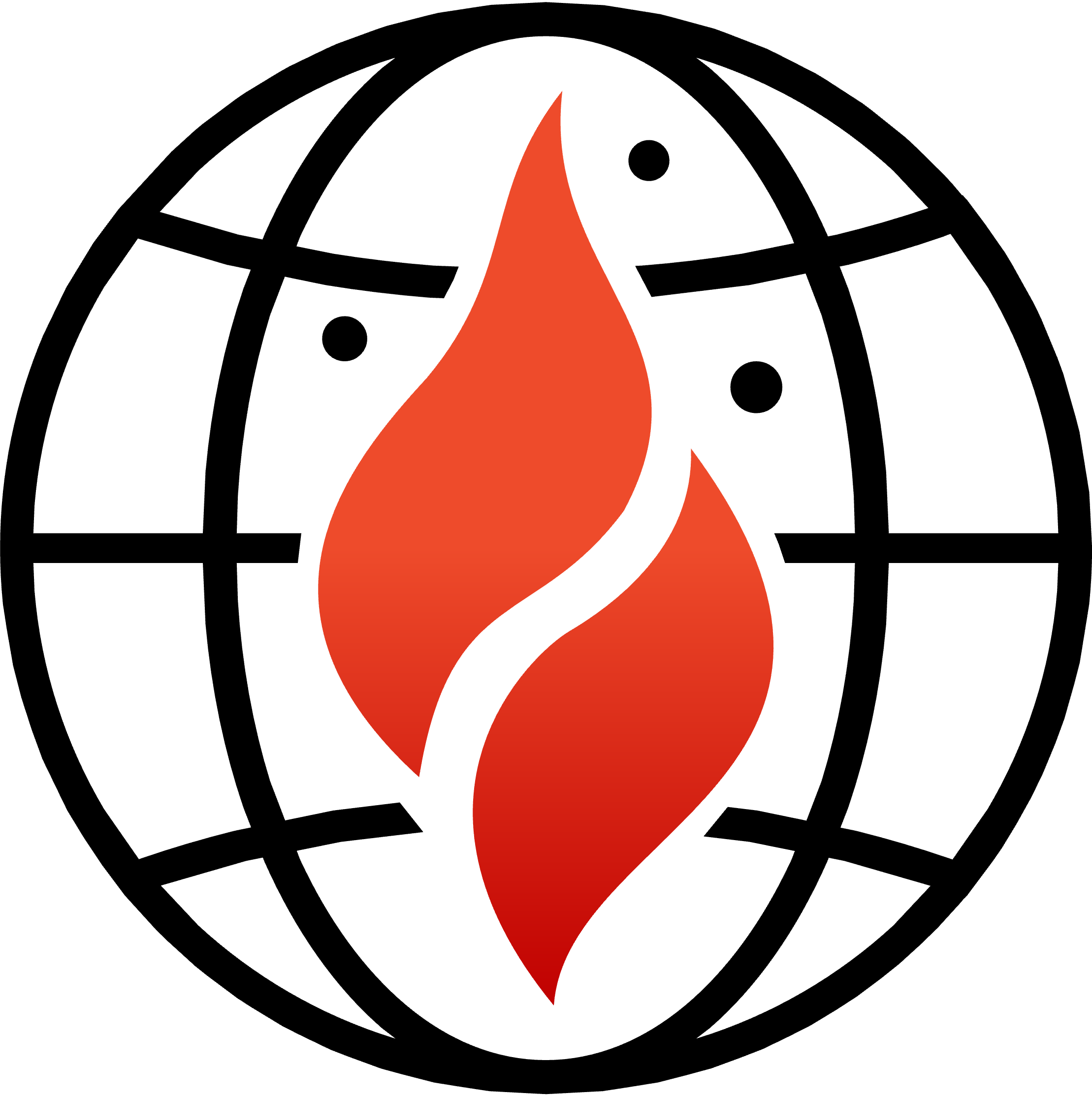}
    \end{minipage}
    \begin{minipage}{0.78\textwidth}
        \centering \uppercase{TerraTorch: The Geospatial\\Foundation Models Toolkit}
    \end{minipage}
}

\makeatletter
\newcommand{\linebreakand}{%
  \end{@IEEEauthorhalign}
  \hfill\mbox{}\par
  \mbox{}\hfill\begin{@IEEEauthorhalign}
}
\makeatother
\author{
\IEEEauthorblockN{
    Carlos Gomes\IEEEauthorrefmark{1}, 
    Benedikt Blumenstiel, 
    Joao Lucas de Sousa Almeida,\\ 
    Pedro Henrique de Oliveira\IEEEauthorrefmark{1}, 
    Paolo Fraccaro, 
    Francesc Marti Escofet\IEEEauthorrefmark{1},\\ 
    Daniela Szwarcman, 
    Naomi Simumba, 
    Romeo Kienzler, 
    Bianca Zadrozny
} \\
\linebreakand
\IEEEauthorblockA{IBM Research}

\thanks{\IEEEauthorrefmark{1} Work done while at IBM Research.}
\thanks{The TerraTorch team can be reached via email at terratorch@ibm.com.}
\pagestyle{plain}
\thanks{© 2024 IEEE. Personal use of this material is permitted. Permission from IEEE must be obtained for all other uses, in any current or future media, including reprinting/republishing this material for advertising or promotional purposes, creating new collective works, for resale or redistribution to servers or lists, or reuse of any copyrighted component of this work in other works.}
}

\maketitle

\begin{abstract}
    TerraTorch is a fine-tuning and benchmarking toolkit for Geospatial Foundation Models built on PyTorch Lightning and tailored for satellite, weather, and climate data. It integrates domain-specific data modules, pre-defined tasks, and a modular model factory that pairs any backbone with diverse decoder heads. These components allow researchers and practitioners to fine-tune supported models in a no-code fashion by simply editing a training configuration. By consolidating best practices for model development and incorporating the automated hyperparameter optimization extension Iterate, TerraTorch reduces the expertise and time required to fine-tune or benchmark models on new Earth Observation use cases. Furthermore, TerraTorch directly integrates with GEO-Bench, allowing for systematic and reproducible benchmarking of Geospatial Foundation Models. TerraTorch is open sourced under Apache~2.0, available at \url{https://github.com/IBM/terratorch}, and can be installed via \texttt{pip~install~terratorch}.
\end{abstract}


\section{Introduction}

Earth Observation (EO), weather, and climate (WxC) data present unique challenges due to their heterogeneity, like varying spectral bands and resolutions, and often require specialized domain knowledge. Machine learning (ML) is essential in EO and WxC as it enables the efficient processing and analysis of large amounts of satellite and weather data, allowing us to derive insights critical for environmental monitoring, disaster response, or resource management~\cite{rolnick2022tackling,jones2023ai}. With the increasing demand for ML models, specialized Geospatial Foundation Models (GeoFMs) such as the Prithvi model family~\cite{prithvi2023,prithviwxc2024,prithvieo22024} or Clay~\cite{clay2024} have emerged. They are pre-trained on millions of satellite images or weather data points, which improves the model accuracy and generalizability in downstream applications while reducing the required compute for fine-tuning.
Nonetheless, fine-tuning GeoFMs demands significant manual work since no generic framework completely automates the process for end users. Additionally, it requires machine learning knowledge that not all researchers have, limiting the effective adoption of GeoFMs.

TerraTorch is designed to address these issues by combining the automation and scalability of PyTorch Lightning~\cite{falcon2019pytorchlightning} with the flexibility of a timm-like model factory~\cite{timm2019} and access to diverse datasets from TorchGeo~\cite{TorchGeo2022}. This synthesis enables users to fine-tune FMs with minimal effort, allowing rapid prototyping and streamlined integration into existing pipelines.

Our framework focuses on modularity, offering an adaptable model architecture where backbones, decoders, and other components can be swapped or extended according to task requirements. It is designed to facilitate the integration of third-party models, datasets, and tasks while avoiding code duplication. By leveraging this modularity, practitioners can address various ML tasks, ranging from scene-level classification to pixel-level segmentation, regression, or downscaling. The package’s consistent interface and generic data modules reduce the manual overhead typically associated with preparing custom datasets for fine-tuning. Additionally, TerraTorch has a built-in hyperparameter optimization (HPO) extension called Iterate to ensure the use of suitable hyperparameters during fine-tuning. Moreover, this module allows us to systematically benchmark and compare GeoFMs on collections like GEO-Bench~\cite{geobench2024}.

In summary, TerraTorch makes the following contributions: (1)~A fine-tuning and inference framework tailored for EO and WxC applications, (2)~well-integrated GeoFMs in a modular architecture supporting a wide range of tasks, and (3)~the Iterate extension for HPO. These features push the boundaries towards our goal of simplifying the adoption of state-of-the-art FMs, ultimately accelerating research and production pipelines for remote sensing applications.

\section{Related work}

PyTorch enabled wider access to high-performance computing by abstracting away CUDA complexity~\cite{cuda2008} and has become the most widely adopted Deep Learning framework in research and industry~\cite{paszke2019pytorch}. Building on PyTorch, Lightning~\cite{falcon2019pytorchlightning} offers high-level abstractions that reduce boilerplate code, much like Keras did for TensorFlow~\cite{chollet2015keras}. 
It enforces a standardized project structure and automatically handles complex training features, including distributed training - all built on PyTorch’s foundation. Lightning integrates seamlessly with popular experiment tracking tools such as TensorBoard~\cite{tensorflow2015tensorboard}, MLflow~\cite{zaharia2018mlflow}, and Weights \& Biases~\cite{wandb2020}. The framework effectively serves as an organizational layer that preserves PyTorch’s flexibility while abstracting away common implementation steps.
TorchGeo~\cite{TorchGeo2022} aligns with this paradigm, adding geospatial data handling through specialized datasets, transforms, and samplers while preserving compatibility with the Lightning ecosystem.
TerraTorch directly inherits from Lightning and TorchGeo modules, providing users with the same ease of use while offering additional functionality.

Timm (PyTorch Image Models)~\cite{timm2019} provides a wide range of state-of-the-art vision models, layers, and utility functions. Users can easily initialize backbones and classification models via the timm package. Similarly, Segmentation Models PyTorch (SMP)~\cite{smp2019} provides multiple models and decoders for segmentation tasks. Therefore, all timm and SMP models are directly available in TerraTorch's model registry.

TerraTorch embraces these ideas by wrapping libraries like Lightning, TorchGeo, timm, and SMP into a single package that serves geospatial, weather, and climate applications. 
While TorchGeo remains at TerraTorch’s core, TorchGeo focuses primarily on geospatial data workflows and provides limited ways to instantiate models. TerraTorch additionally incorporates automated HPO (Iterate), benchmarking support for dataset collections, and the flexible integration of external model repositories. 

\section{Design}

TerraTorch is a comprehensive framework tailored for geospatial applications, leveraging the Lightning library with specific adaptations for our domain. Its design emphasizes extensibility, user-friendliness, and compatibility with tools such as TorchGeo. We leverage Lightning's comprehensive automation tools, such as GPU scaling, checkpointing, and logging, while ensuring the integration of TorchGeo datasets and pre-trained tried-party models. 

Multiple abstraction levels in TerraTorch allow users with varying technical expertise to approach fine-tuning and benchmarking in different ways: no-code through YAML configuration files and a command-line interface or via Python scripts/notebooks that integrate TerraTorch’s components for custom workflows. For users requiring only specific elements, TerraTorch supports standalone initialization of models for use in external pipelines. This modular design ensures flexibility across diverse applications.

A key element of TerraTorch’s extensibility is its inheritance from TorchGeo datasets and tasks, as well as its modular model architecture definition with backbones, necks, decoders, and heads. This ensures broad compatibility with TorchGeo datasets and third-party backbones, e.g., from timm or SMP. More importantly, TerraTorch integrates a range of GeoFMs, like Prithvi or Clay, and ensures straightforward integration of newly released models. All available GeoFMs undergo extensive testing to ensure a functional end-to-end pipeline. 

\begin{figure}[tbh]
	\centering
	\includegraphics[width=\linewidth]{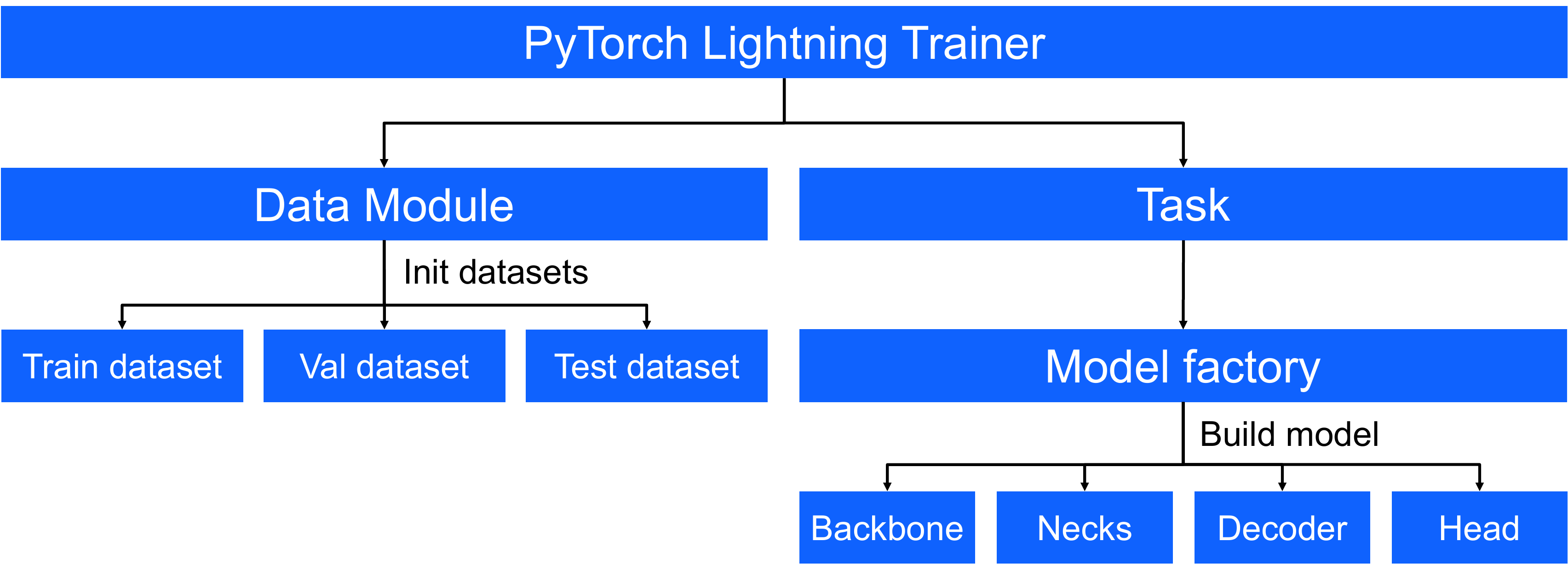}
	\caption{Overview of the TerraTorch components and structure.}
    \label{fig:overview}
\end{figure}

As illustrated in Figure~\ref{fig:overview}, TerraTorch follows a framework similar to Lightning and TorchGeo. The Lightning trainer supervises training, validation, and testing, as well as taking care of optimizers and low-level GPU management. TerraTorch tasks encapsulate the training and evaluation logic, defining data flow and metrics. Data modules handle the initialization of datasets for different splits, which include loading the samples and applying the necessary transformations. The model factory builds the model architecture using a modular approach by pairing a chosen backbone with necks, a task-specific decoder, and the head. Incorporating EO- and WxC-specific functionalities on various levels—including spectral channel handling for the model inputs and best practices for data preprocessing with generic dataset classes—simplifies model fine-tuning for users. 
Furthermore, we provide tutorials and notebooks to flatten the learning curve for new users, helping them quickly adapt to the framework’s functionalities. These elements collectively guide users through the development process, from data ingestion to the final model analysis.

The extension Iterate\footnote{Iterate is available at \url{https://github.com/IBM/terratorch-iterate}.} performs Bayesian HPO to improve the model performance. By specifying the default and tunable parameters in a single YAML configuration, practitioners can automate HPO with minimal overhead and ensure the fine-tuning of FMs with suitable hyperparameters. Iterate dramatically reduces manual work and enables consistent benchmarking. Once the model is fine-tuned, it can be integrated into a geospatial application using TerraTorch's inference tool.


\section{Implementation}

\begin{table*}[htb]
\centering
\small
\setlength{\tabcolsep}{3.5pt}
\renewcommand{\arraystretch}{1.2}
\caption{Overview of selected EO backbones available in TerraTorch and their pre-training characteristics.}
\label{tab:backbones}
\begin{tabularx}{\textwidth}{lccccccX}
\toprule
Model & Architecture & \# Params. & Technique & Sensors & Resolution & \# Samples & Features \\
\midrule
MOCOv2~\cite{ssl4eo2023} & ResNet50 & 25M & Contrastive & \mbox{Sentinel-2} & 10m & 1M & Multi-seasonal contrastive learning \\
DINO~\cite{ssl4eo2023} & ResNet50 & 25M & Contrastive & \mbox{Sentinel-2} & 10m & 1M & Multi-seasonal contrastive learning \\
DeCUR~\cite{decur2024} & ResNet50 & 25M & Contrastive & \mbox{S1, S2} & 10m & 1M & Radar-optical contrastive learning \\
Satlas~\cite{satlas2023} & Swin & 100M & Supervised & various & 10m & NA & Multi-task pre-training \\ 
ScaleMAE~\cite{scalemae2023} & ViT & 300M & MAE & RGB & 0.1-30m & 360k & Resolution-based pos. embedding \\ 
DOFA~\cite{dofa2024} & ViT & 300M & MAE & various & 1-30m & 8M & Dynamic wave-lengths encoding \\
Clay v1~\cite{clay2024} & ViT & 100M & MAE \& DINO & various & 1-500m & 70M & Location and temporal encodings, dynamic wave-lengths encoding \\ 
Prithvi-EO-1.0~\cite{prithvi2023} & ViT & 100M & MAE & HLS & 30m & 250k & Multi-temporal inputs\\ 
Prithvi-EO-2.0~\cite{prithvieo22024} & ViT & 300M, 600M & MAE & HLS & 30m & 4.2M & Location and temporal encodings, multi-temporal inputs  \\
\bottomrule
\end{tabularx}
\end{table*}

TerraTorch organizes its core functionality into four main components. Tasks define the training, evaluation and inference logic. The model factory integrates various backbones, necks, decoders, and heads. Generic datasets simplify data ingestion and preprocessing for different tasks. Finally, Iterate offers hyperparameter optimization and repeated runs, allowing users to automate parameter tuning for improved model performance and benchmarking.

\subsection{Tasks}

TerraTorch leverages Lightning to define training, validation, test, and inference steps in a structured manner. Each task manages the training loss, metric computation, and optional visualizations of predictions for monitoring model performance. Building on TorchGeo’s tasks, TerraTorch adds features like support for multi-modal inputs or shifting-window inference.
TerraTorch currently supports semantic segmentation, pixel-wise regression, classification, object detection, and downscaling. The inference task can be used in production pipelines to directly deploy fine-tuned GeoFMs. Models are typically created using TerraTorch’s model factory, but users can also provide custom PyTorch models for full flexibility. 

\subsection{Model factory}

In TerraTorch, users can modularly combine backbones and decoders for enhanced flexibility. This functionality is realized through the \texttt{EncoderDecoderFactory}, responsible for initializing both the backbone and decoder. Furthermore, necks facilitate the conversion of layer-wise backbone outputs into the required decoder inputs. For instance, users can choose layer indices for hierarchical decoders such as UPerNet~\cite{upernet2018} or transform the 1D patched outputs into a 2D grid. Finally, a head is added to generate the prediction.

Users can find the available backbones in the \texttt{BACKBONE\_REGISTRY}, and we highlight some GeoFMs in Table~\ref{tab:backbones}. The supported models include DeCUR~\cite{decur2024}, which is trained with contrastive learning, Clay~v1~\cite{clay2024} with its dynamic wavelength encoding, and the multi-temporal Prithvi models~\cite{prithvi2023,prithvieo22024}.
Additionally, backbones from timm and SMP are available. Future FMs can be easily added to TerraTorch by registering a model to the backbone registry.

TerraTorch includes a range of decoders like UPerNet or fully convolutional networks (FCN), which are helpful for pixel-wise tasks. Because low-level features are often not required for image-level tasks, the \texttt{IdentityDecoder} is suited for tasks like classification. The \texttt{ObjectDetectionTask} uses decoders from Torchvision~\cite{torchvision2016}, such as Faster-RCNN~\cite{fasterrcnn2016}.

Users can also pass PyTorch modules for the backbone or decoder instead of choosing one from the registry. The model factory also supports methods such as LoRA~\cite{lora2021} for parameter-efficient fine-tuning through the PEFT~\cite{peft2022} library.

\subsection{Datasets}

Lightning and TorchGeo typically rely on custom data modules and dataset classes, which can be difficult to understand for newcomers. To simplify this process, TerraTorch provides generic data modules that support no-code fine-tuning for tasks like classification, semantic segmentation, and pixel-wise regression. For classification, users can adopt a standard \texttt{ImageFolder} structure, where each split (train, val, test) is stored in separate folders, and samples are organized in subfolders by class. So, all training samples of \textit{class i} are stored in \texttt{dataset/train/class~i/}. Pixel-wise tasks allow greater flexibility: images and labels can be stored together or separately, and splits can be managed by folder structures or split files. Simple parameters such as \texttt{image\_grep} allow users to handle many dataset formats, reducing potential pre-processing work. In addition, generic datasets can process temporal inputs, which must be stacked as bands in single TIF files per time series. The datasets automatically unstack the temporal dimension based on the user-defined \texttt{dataset\_bands}. Natively, TerraTorch can consume any dataset and data module from TorchGeo, and additional ones can be easily added. Of particular importance is the integration of all GEO-Bench~\cite{geobench2024} datasets, which makes TerraTorch a strong candidate for reliably benchmarking GeoFMs. 

\subsection{TerraTorch Iterate}

TerraTorch provides a tool called Iterate, which provides a framework around TerraTorch to perform HPO and repeated experiments. It leverages MLFlow~\cite{zaharia2018mlflow} for experiment logging, Optuna~\cite{akiba2019optuna} for Bayesian HPO, and optionally Ray~Tune~\cite{moritz2018ray} for parallelization. Users can perform hyperparameter optimization across multiple tasks (e.g., different datasets or backbones) by defining a benchmark in a single YAML configuration file. 
Default parameters and task-specific parameters are combined to form the training configuration, while an \texttt{optimization\_space} block controls which hyperparameters to explore. Instead of using random search or grid search, Optuna uses Bayesian optimization to approximate good hyperparameters with fewer trials. The trials can be parallelized on multiple GPUs in a Ray cluster using Ray~Tune.
After finding a satisfactory configuration for each task, Iterate allows users to rerun experiments multiple times with different random seeds for robust comparisons, following the established practice in benchmarking~\cite{geobench2024,prithvieo22024}.

\section{Experiments}

We present two examples of potential downstream tasks to demonstrate TerraTorch's capabilities. Table~\ref{tab:results} provides a comparison of four EO FMs for the datasets Sen1Floods11~\cite{sen1floods112020} and BurnScars\footnote{We used the new splits for BurnScars from~\cite{prithvieo22024} as the original dataset does not include a test set.}~\cite{burnscars2023}. We report the IoU for the classes \textit{water} and \textit{burned} as well as the mean IoU. All results are created using TerraTorch Iterate, which optimized the validation loss by adjusting the learning rate and weight decay over 10 runs. 
Table~\ref{tab:results} presents the test set results of the best runs based on the validation loss. All three models process 13 channels from S2L1C for Sen1Floods11 and six Harmonized-Landsat-Sentinel-2 (HLS) channels for BurnScars, using a batch size of 16 and an input size of 512 with random rotations and flips. 
Because the Prithvi models are pre-trained on six HLS channels, TerraTorch automatically updates the patch embeddings for S2 data by coping the pre-trained weights for matching channels while randomly initializing unseen ones. Clay and Prithvi-EO-2.0 additionally process metadata (time and location).
We use a UPerNet decoder~\cite{upernet2018} for all models with four input layers.
For the Vision Transformer models~\cite{vit2020}, we apply a TerraTorch neck to upscale the intermediate outputs.
We used the \texttt{AdamW} optimizer with an \texttt{ReduceLROnPlateau} scheduler and trained for 100 epochs with an early stopping after 20 epochs of no improvement. The Iterate runs took between 4h (DeCUR) and 46h (Clay~v1) on an NVIDIA-A100-80G.

\begin{table}[tbh]
	\centering
    \small
    \setlength{\tabcolsep}{1pt}
    \renewcommand{\arraystretch}{1.2}
	\caption{Fine-tuning results on the test splits.}
    \label{tab:results}
	\begin{tabularx}{\linewidth}{lCCCC}
		\toprule
         & \multicolumn{2}{c}{Sen1Floods11} & \multicolumn{2}{c}{BurnScars} \\
        Model & $IoU_{water}$ & $mIoU$ & $IoU_{burned}$ & $mIoU$ \\
        \midrule
        DeCUR & 82.17 & 89.75 & 84.94 & 91.53 \\
        Clay v1 & \textbf{82.43} & \textbf{89.90} & 87.68 & 93.09 \\
        Prithvi-EO-1.0-100M & 80.19 & 88.62 & 80.97 & 89.32 \\
        Prithvi-EO-2.0-300M-TL & 80.97 & 89.06 & \textbf{88.17} & \textbf{93.37} \\
        \bottomrule
	\end{tabularx}
\end{table}

These results demonstrate how both model architecture and pre-training data influence downstream performance. DeCUR and Clay~v1, pre-trained on Sentinel-2, show strong results on Sen1Floods11.
By contrast, Prithvi-EO-2.0-300M-TL performs best on BurnScars, which relies on HLS data and thus matches the Prithvi pre-training data. Prithvi-EO-1.0-100M serves as a baseline and is outperformed by the other models trained on global data. The contrastive learning of DeCUR, as well as the patch size 8 of Clay v1, might additionally explain the better performance compared to Prithvi-EO-1.0. Overall, these findings highlight the importance of having a variety of pre-trained models available in TerraTorch since no single architecture perfectly covers all use cases.

\begin{figure}[htb]
	\centering
     \begin{subfigure}{\linewidth}
         \centering
         \includegraphics[width=\linewidth]{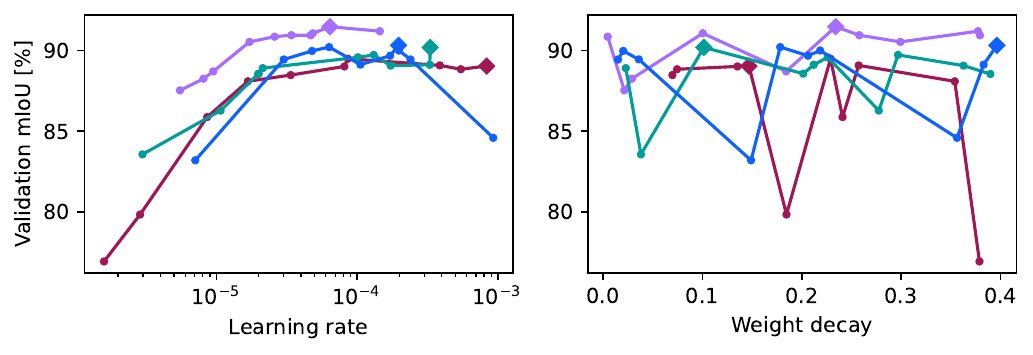}
         \caption{Sen1Floods11}
         \label{fig:hpo_sen1floods11}
         \vspace{2mm}
     \end{subfigure}     
     \begin{subfigure}{\linewidth}
         \centering
         \includegraphics[width=\linewidth]{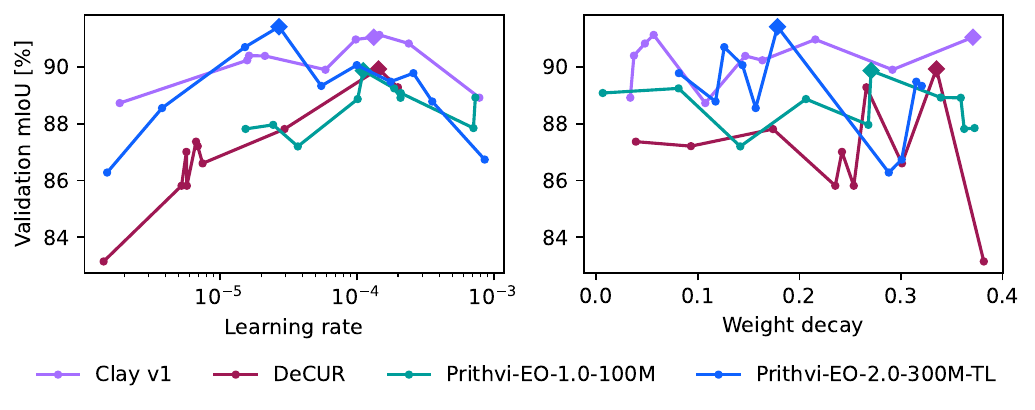}
         \caption{BurnScars}
         \label{fig:hpo_burn_scars}
     \end{subfigure}
	\caption{Effect of hyperparameters on the validation mIoU per dataset. The best runs are marked with a diamond.}
    \label{fig:hpo}
\end{figure}

Figure~\ref{fig:hpo} illustrates all trials by Iterate and how different learning rates and weight decay values impacted the validation mIoU. The best run is selected based on the lowest validation loss, which is highlighted with a diamond. Notably, learning rates around 1e-4 tend to yield stable training and higher scores, while either very low or high rates often lead to suboptimal performance. While most researchers are using default values in this suitable range, small changes using HPO easily improve the model performance by 2pp.
In comparison, weight decay plays a less critical role as no trend is visible, and the variations are clearly influenced by the learning rate. 
This analysis highlights the need for systematic HPO to unlock the full potential of ML and how it can be used to identify the importance of parameters.

\section{Future work}

The upcoming development of TerraTorch will expand its functionality in handling multi-modality and multi-sensor data and emphasize the integration of openEO~\cite{openeo2017} for data streaming. Our roadmap includes introducing new task types, like neural compression and image generation. However, our primary focus remains on integrating new FMs for EO and WxC to ensure ease of use. These steps aim to create a more comprehensive framework for various EO applications. We actively encourage users to contribute to TerraTorch and provide us with feedback about functionality and user experience.

\section{Conclusion}

TerraTorch provides an end-to-end solution for fine-tuning ML models for EO tasks, covering data handling, training workflows, Bayesian HPO, and model benchmarking. By integrating PyTorch Lightning, TorchGeo, and other libraries, TerraTorch reduces the complexity of EO and WxC model development and accelerates research and operational pipelines. Its modular, extensible design addresses the growing need for reliable, domain-specific tools in remote sensing, weather, and climate analytics.

\section{Acknowledgement}

We thank all TerraTorch contributors for their invaluable support. Their efforts in code development and testing ensure that TerraTorch remains an up-to-date and user-friendly framework. We also extend our gratitude to the PyTorch Lightning, TorchGeo, and broader open-source communities whose work has paved the way for TerraTorch’s development.

\small
\bibliographystyle{IEEEtranN}
\bibliography{references}

\end{document}